\documentclass{jmlr}
\pdfoutput=1

\jmlrworkshop{ICML 2018 AutoML Workshop}

\usepackage{bm}
\usepackage{graphicx}
\usepackage{hyperref}       
\usepackage{url}            
\usepackage{wrapfig}

\usepackage[font=small,labelfont=bf]{caption}

\title[Joint Neural Architecture and Hyperparameter Search]{Towards Automated Deep Learning: Efficient \\ Joint Neural Architecture and Hyperparameter Search}

  \author{\Name{Arber Zela} \Email{zelaa@cs.uni-freiburg.de}\\
   \Name{Aaron Klein} \Email{kleinaa@cs.uni-freiburg.de}\\
   \Name{Stefan Falkner} \Email{sfalkner@cs.uni-freiburg.de}\\
   \Name{Frank Hutter} \Email{fh@cs.uni-freiburg.de}\\
   \addr Department of Computer Science, University of Freiburg}

\begin{document}

\maketitle

\begin{abstract}
 While existing work on neural architecture search (NAS) tunes hyperparameters in a separate post-processing step, we demonstrate that architectural choices and other hyperparameter settings interact in a way that can render this separation suboptimal.
  Likewise, we demonstrate that the common practice of using very few epochs during the main NAS and much larger numbers of epochs during a post-processing step is inefficient due to little correlation in the relative rankings for these two training regimes.
  To combat both of these problems, we propose to use a recent combination of Bayesian optimization and Hyperband for efficient joint neural architecture and hyperparameter search.
\end{abstract}

\section{Introduction}


Before the rise of deep learning and its success in end-to-end feature learning, manual feature engineering was arguably one of the most important steps in the machine learning workflow, but also very time-consuming and tedious.
With an abundance of choices in designing the architecture of deep neural networks, manual feature engineering has nowadays to a certain degree been replaced by manual tuning of architectures.
Recent work on \emph{neural architecture search (NAS)}~\citep{baker-iclr17, enas, zoph-iclr17, real-icml17a, zoph-arXiv18,real-arxiv18a,liu-arXiv17,cai-aaai18, elsken-arXiv18} automates this choice of network architecture.
On some benchmarks, these NAS methods have indeed led to new state-of-the-art performance~\citep{zoph-iclr17,real-arxiv18a}, although only at extreme computational costs on the order of 800 GPUs for two weeks.


Many prominent NAS methods~\citep{zoph-iclr17,real-icml17a,zoph-arXiv18,real-arxiv18a} follow a two step process, which we argue is inefficient. During their main architecture search phase, they evaluate architectures with a fixed set of hyperparameters and a relatively small number of epochs (e.g., 20), and only after the search has finished, they optimize hyperparameters for the end result and evaluate it with a large number of epochs (e.g., 600). This process is suboptimal for various reasons:
\begin{itemize} 
  \item The resulting approach is not an anytime approach and does not satisfy the requirement of an automated machine learning (AutoML) system~\citep{feurer-nips2015a} to make predictions after a given time budget.
  \item The sudden jump from a small budget of 20 to a large budget of 600 epochs leads to little correlation between the relative ranks on the small and the large budgets, potentially rendering most of the expensive optimization at the small budget void. 
\end{itemize}  
To combat these problems, we propose to use an approach for joint neural architecture and hyperparameter search that is anytime and gradually increases the computational budget for the best fraction of networks at lower compute budgets in order to yield a far more computationally efficient optimization procedure.  
Specifically, our contributions are as follows:
\begin{itemize} 
  \item We show how to use a recent combination~\citep{falkner-iclr2018a} of Bayesian optimization~\citep{shahriari-ieee16a} and Hyperband~\citep{li-iclr17} to perform efficient joint neural architecture and hyperparameter search. 
  \item We demostrate the weak correlation between performance after short and long training budgets, which potentially affects both architecture and hyperparameter choices when optimized with these two budgets, and show how to avoid this effect by incrementally increasing the budget during the optimization process.
  \item We show that for a limited training runtime of 3 hours we can achieve competitive performance on CIFAR-10 if we optimize the hyperparameters and architecture jointly.
    
  \end{itemize}




\section{Related Work}

\citet{melis-iclr18} showed that a well-tuned LSTM~\citep{hochreiter1997long} with the right training pipeline was able to outperform a recurrent cell found by neural architecture search~\citep{zoph-iclr17} on the Penn Treebank dataset. This underlines the effect hyperparameters can play in practice.

One of the most successful methods to optimize the hyperparameters of deep neural networks is Bayesian optimization \citep{snoek-nips12a, hutter-lion11a, bergstra-nips11a}.
Since each function evaluation requires to train and evaluate a deep neural network, more advanced Bayesian optimization methods try to speed up the optimization process by exploiting available fidelities of the objective function, such as learning curves~\citep{domhan-ijcai15, klein-iclr17, falkner-iclr2018a} or dataset subsets~\citep{klein-ejs17}.
While most Bayesian optimization methods only focus on a few continuous hyperparameters, \citet{bergstra-icml13a} and \citet{mendoza-automl16a} optimized architecture and hyperparameters jointly and achieved, at this time, state-of-the-art performance for shallow convolutional neural networks and feed forward neural networks, respectively.
 
There are many related methods for neural architecture search, for example based on reinforcement learning~\citep{zoph-iclr17,zoph-arXiv18} and evolutionary algorithms~\citep{real-icml17a,liu-arXiv17,real-arxiv18a}. All of these methods focus only on the neural architecture, keeping hyperparameter fixed during the search (and optimizing them in a post-hoc step).

\section{Efficient Joint Hyperparameter Optimization and Architecture Search}
\label{sec:bohb}

In this section we discuss how to cast neural architecture search as a hyperparameter optimization problem and tackle it together with the standard hyperparameter optimization problem. We also discuss a method for efficiently searching in this joint space.


\subsection{Neural Architecture Search as Hyperparameter Optimization}

While the NAS literature casts the architecture search problem as very different from hyperparameter optimization, we argue that most NAS search spaces can be written as hyperparameter optimization search spaces (using the standard concepts of categorical and conditional hyperparameters). Indeed, this fact also enables the use of standard evolutionary algorithms in the literature~\cite{real-icml17a, real-arxiv18a}.

As an example, take the parameterization of a convolutional Normal Cell introduced by \citet{zoph-arXiv18}. Each cell receives as inputs two previous hidden states (the feature maps of two cells in previous layers, or the input image directly), 
and outputs a new hidden state. The Normal Cell is composed of $B$ (by default, $B=5$) blocks
%
%
and the $k-th$ block consists of 5 categorical choices: 
select a first hidden state (out of the cell's 2 inputs and the output hidden states of blocks $1, \ldots, k-1$); select a second hidden state (out of the same domain as the first choice); select an operation (out of 13 operations including several types of convolutions, pooling operations and the identity; see \citet{zoph-arXiv18} for the full list) to apply to the first hidden state; select an operation to apply to the second hidden state (out of the same 13 operations); and select a method (element-wise addition or concatenation) to combine the outputs of these operations to create a new hidden state (which is added to the existing set of hidden states).
In the end, all the unused output hidden states from the cell's $B$ blocks are concatenated together to form the final cell output (there are no free choices in this step).
Therefore, in total, this search space is fully specified by $5B$ categorical hyperparameters.
%

Casting NAS as a hyperparameter optimization problem with categorical hyperparameters
immediately suggests the possibility of joint architecture and hyperparameter search by just extending the hyperparameter space for the NAS part by the standard hyperparameters to be tuned. Likewise, it would be possible to sample hyperparameters in an RL approach or optimize them via genetic algorithms alongside the neural architecture.

\subsection{Bayesian Optimization Hyperband}

To efficiently optimize in the joint space of architectures and hyperparameters, we use BOHB~\citep{falkner-iclr2018a}, a recent combination of Bayesian optimization~\citep{shahriari-ieee16a} and Hyperband~\citep{li-iclr17}. Due to space constraints, we refer the reader to \cite{falkner-iclr2018a} and \cite{li-iclr17} for full details on these methods and only provide the basics here. 

Like Hyperband, BOHB uses evaluations on different budgets to accelerate the optimization by exploiting knowledge gained on cheap, lowfidelity observations. It dynamically allocates more resources to promising configurations by repeatedly invoking the Successive Halving~\citep{jamieson-aistats16} subroutine. 
Successive Halving evaluates a large number of configurations using a small minimum budget $b_{min}$ and continues to evaluate the best $\eta^{-1}$ (by default best-performing third) of these with the next budget $\eta\cdot b_{min}$. This is repeated until reaching a maximum budget $b_{max}$. As an example, consider a budget based on the number of epochs a neural network is trained for; $b_{min}$ could, e.g., be 10, and $b_{max}$ 270.

Like Bayesian optimization, BOHB uses a probabilistic model to sample promising configurations rather than selecting these uniformly at random as done in Hyperband. BOHB uses multivariate Kernel Density Estimators (KDEs) over the input configuration space to better model interactions between parameters and returns a new sample with the highest expected improvement (EI).

%

\subsection{Joint Architecture and Hyperparameter Search Space}
We picked a multiple-branch ResNet architecture as the basis for our search space (see \cite{gastaldi-iclr17}). The first layer is a 3x3 convolution followed by 3 main blocks, a 8x8 average pooling and a fully connected layer in the end for discriminating between classes. We parametrized the number of filters $Filters_0$ for the first convolution and the number of residual blocks $ResBlocks_j$, branches $ResBranches_j$ and filters (determined by a widening factor $WidenFactor_j$) within each main block $j$. However, we kept a fixed structure (\textsf{ReLU-Conv3x3-BN-ReLU-Conv3x3-BN-Mul}) for each residual branch. Each sampled architecture configuration $Net_i$, is thus defined by these 10 architectural choices as:

\begin{equation}
\begin{split}
	& Net_i = stack_{j=1}^3 \lbrace MainBlock_j \rbrace ,	 \\
	& MainBlock_j = stack_{k=1}^{ResBlocks_j} \lbrace ResBlock_k \rbrace ,	 \\
	& ResBlock_k = add_{l=1}^{ResBranches_j} \lbrace ResBranch_l \rbrace	, \\
	& Filters_j = round(WidenFactor_j \cdot Filters_{j-1}	), \\
\end{split}
\end{equation}
where $stack_{i=1}^{N} \lbrace motif_i \rbrace$ applies motifs sequentially and $add_{i=1}^{N} \lbrace motif_i \rbrace$ adds their outputs element-wise. For example, if $\mathcal{F}_1(\cdot)$ and $\mathcal{F}_2(\cdot)$ denote transformations applied to input $x$ by $ResBlock_1$ and $ResBlock_2$ respectivelly, $stack \lbrace ResBlock_1, ResBlock_2 \rbrace$ yields $\mathcal{F}_2(\mathcal{F}_1(x))$ as output. In the case of $ResBranch_1$ and $ResBranch_2$, $add \lbrace ResBranch_1, ResBranch_2 \rbrace$ would result in $x + \mathcal{F}_1(x) + \mathcal{F}_2(x)$. One could combine these different operations and motifs recursively in order to search for more complex architectures as the system evolves, but we restrict ourselves to this 10-dimensional architecture space due to a limited compute budget.
We leave for future work the design of a more generic search space which does not limit the network information into a fixed length vector.

We optimized this network using the regularization methods of CutOut~\citep{cutout}, MixUp~\citep{mixup}, Shake-Shake~\citep{gastaldi-iclr17}, and Shake-Drop~\citep{yamada2018shakedrop}\footnote{For residual blocks containing $b>1$ residual branches, Shake-Shake scales the feature maps from branch $i$ by a random factor $a_i$, such that $\sum_{i}^{b} a_i = 1$. Only in the case of $b=1$ we apply ShakeDrop instead.}, as well as the optimization algorithm of stochastic gradient descent with restarts (SGDR; \citet{loshchilov-ICLR17SGDR}).
As hyperparameters, we tuned initial learning rate, batch size, momentum, $L_2$ regularization, CutOut length, and the $\alpha$ parameter for MixUp, as well as the death rate for ShakeDrop~\citep{yamada2018shakedrop}. Table \ref{tab:search_space} in Appendix \ref{sec:search_space} summarizes these 10 architectural choices and 7 hyperparameters and provides the specific ranges we used.

\section{Experiments}

We now empirically study the results of applying BOHB to our joint architecture and hyperparameter search on CIFAR-10. Furthermore, we investigate the characteristics of our search space accross different budgets, including the importance of architectural choices and hyperparameters on performance.

\subsection{NAS Under Training Time Constraints}

In our experiments, due to limited computational resources, and in order to explore a training-time-constrained NAS paradigm, we do not allow individual training times as large as those required to obtain the state-of-the-art performance, but limited the training time\footnote{We only measured the actual time spend in forward and backward passes during training and exclude any preprocessing and validation overhead.} to a maximum of three hours for each sampled configuration (the minimal budget we consider is 400 seconds).
We used the default settings of BOHB, which results in budgets that differ by a multiplicative factor of $\eta=3$; with a maximum budget of 3h, this yielded budgets of 400s, 1200s, 1h and 3h.

\subsection{Results}

\begin{wrapfigure}{r}{8cm}
\vspace*{-1cm}
\centering
\includegraphics[width=0.9\linewidth]{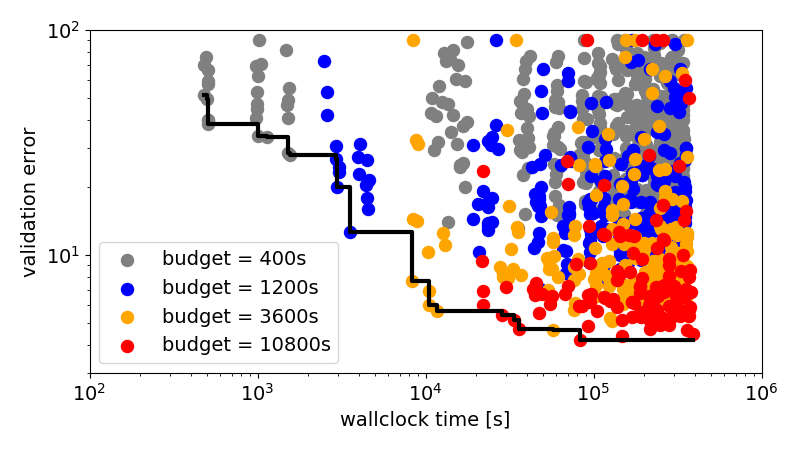}
\caption{Validation error of all configurations evaluated on the different budgets during the whole optimization procedure. The best performing configuration (incumbent) as a function of time is visualized by the black line.}
\label{fig:stepwise_optimization}
\end{wrapfigure}

We ran BOHB for an equivalent of 256 evaluations on the full budget of 3h (i.e., a total of 32 GPU days) and show the results in Figure \ref{fig:stepwise_optimization}. This provides the best performance found as a function of time, as well as the performance of all trained networks, which shows that our search did not only cover the good regions of the space, but also explored sufficiently.
Surprisingly, the final performance reached within this 3h budget was as low as 3.18\% test error; as Table \ref{tab:config_budget} 
shows, this is lower than what can be obtained with several different standard architectures that were also part of BOHB's search space, trained for 3h using the same optimization pipeline and hyperparameters. Training details for the other architectures can be found in Appendix \ref{sec:training_details}. This demonstrates the benefit of optimizing both architecture and hyperparameters.

From Table \ref{tab:config_budget}, we also note that, amongst the three different sizes of multi-branch residual architecture regularized with Shake-Shake, within the 3h time budget the medium-sized architecture (26 2x64d) performed best, not the largest one (26 2x96d). This is in stark contrast to comparisons with a large budget of 1800 epochs, for which the largest architecture performs much better~\citep{gastaldi-iclr17,cutout}. We believe this effect is due to the strong stochasticity introduced by Shake-Shake regularization, which prevents obtaining very good results quickly. Indeed, in a follow-up experiment with a WRN architecture~\citep{zagoruy-corr16} outside of BOHB's design space, which was incompatible with Shake-Shake regularization, we obtained even slightly better results for a time budget of 3h (but worse results for large budgets). This demonstrates an interaction effect between the efficacy of the regularization method used and the available time budget. 


\subsection{Analysis}

We now analyse the characteristics of our search space on the different budgets.
By design, BOHB optimizes the validation performance for each budget starting with the smallest one and moving to the next larger one as soon as enough evaluations have finished  successfully. 
Consequently, the error rates for 400s have been extensively optimized, with coverage of the search space gradually decreasing for larger budgets.

\begin{table}[t]
\vspace*{-0.3cm}
\footnotesize
	\begin{minipage}[t]{0.36\linewidth}
		\captionof{table}{Spearman rank correlation coefficients of the validation errors between different budgets. The correlation is high between every budget and the next larger one, but degrades quickly beyond that.}
		\label{tab:correlations}
		\centering
		\begin{tabular}{l|ccc}
		        &  $1200$s & $1$h  & $3$h \\
		\hline
		$400$s  &   0.87   & 0.31  & 0.05\\
	 	$1200$s &          & 0.88  & 0.64\\
	 	$1$h    &          &       & 0.86\\
		\end{tabular}
	\end{minipage}
	\hfill
	\begin{minipage}[t]{0.6\linewidth}
	\captionof{table}{Comparison of test performance between manually-constructed architectures and the network configuration found by BOHB when trained for a 3 hour budget. All networks used the same optimization pipeline and hyperparameters.}
	\label{tab:config_budget}
	\centering
	\begin{tabular}{l|c|c}
	Network & \textbf{Params} & \textbf{Test error (\%)} \\
	\hline
	ResNet-18 & 11.2M & $3.34 \pm 0.11$ \\ [1pt]
	Shake-Shake 26 2x32d & 2.9M & $3.91 \pm 0.09$ \\[1pt]
	Shake-Shake 26 2x64d & 11.7M & $3.38 \pm 0.07$ \\[1pt]
	Shake-Shake 26 2x96d & 26.2M & $4.22 \pm 0.06$ \\[1pt]
	Ours & 27.6M & $\bm{3.18 \pm 0.16}$ \\[6pt]
	
	\end{tabular}
	\end{minipage}
\end{table}

First, we studied the rank correlation of the final validation error of all configurations that were trained on any particular pair of budgets\footnote{We want to emphasize that most of the evaluations BOHB performed with a 3h budget are for configurations that perform well on smaller budgets, and thus our samples are skewed towards good configurations.} (see Appendix \ref{sec:correlation_budgets} for all possible combinations of budgets).
The results, summarized in Table \ref{tab:correlations}, clearly show that the relative performance of two configurations generalizes to \emph{somewhat} longer training times (e.g., correlations of roughly 0.87 for 3-fold increases of training time), but that it quickly degrades with larger differences in budget and almost vanishes when jumping directly from 400s to 3h. This means in particular that the ranking of the top configurations cannot be deduced from the ranking of much shorter runs only, which is a common practice in the current neural architecture search literature.

We now study the characteristics of the search space using functional analysis of variance (fANOVA; \citet{hutter-icml14a}).
This method allows us to quantify the importance of architecture choices and hyperparameters based on the whole search space by marginalizing performances over all possible values that other hyperparameters could take based on a model fit on the observations.
The importance of a single choice, or a combination of any number of choices, is quantified by the percentage of the performance variation that is explained by only this choice/these choices.
To improve the quality of the analysis, we focus on the results after 400s up to the 1h budget, since enough evaluations have finished on these allowing us to draw meaningful conclusions for these budgets.

\begin{figure}[ht]
	\centering
	\includegraphics[width=0.30\linewidth]{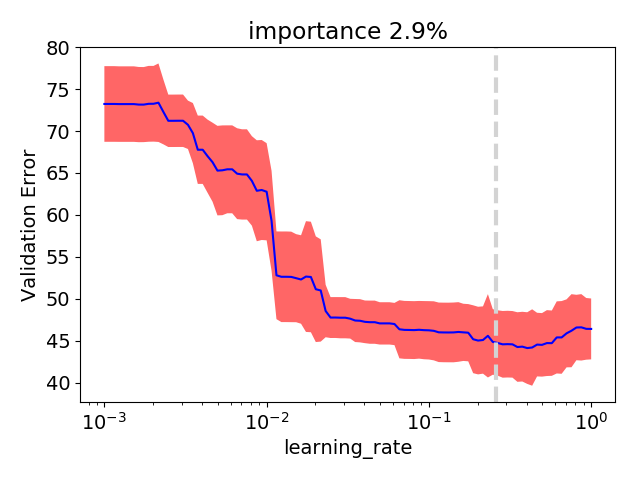}
	\includegraphics[width=0.30\linewidth]{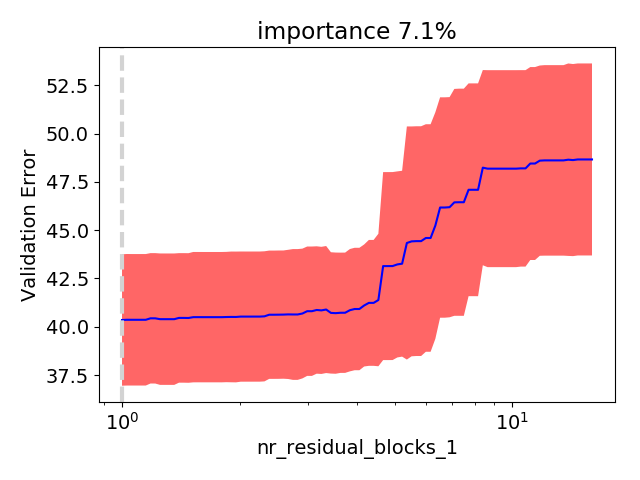}
	\includegraphics[width=0.30\linewidth]{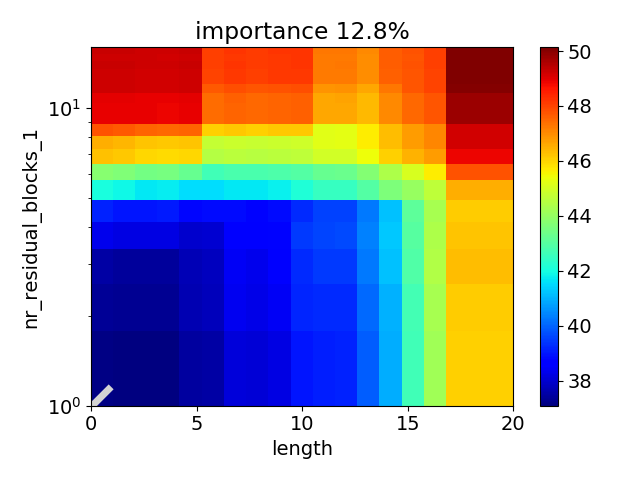}
	
	\includegraphics[width=0.30\linewidth]{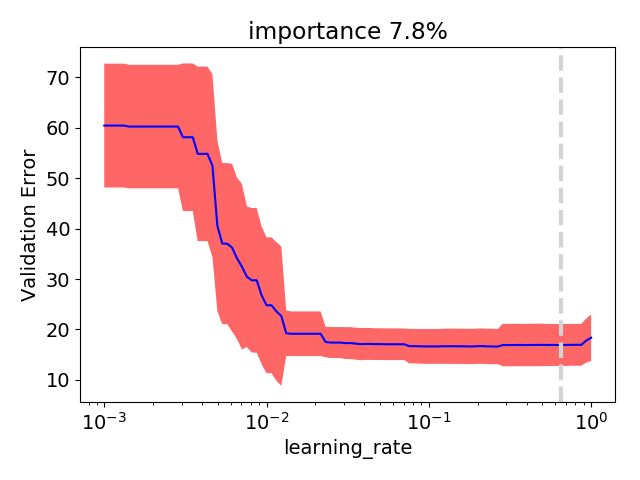}
	\includegraphics[width=0.30\linewidth]{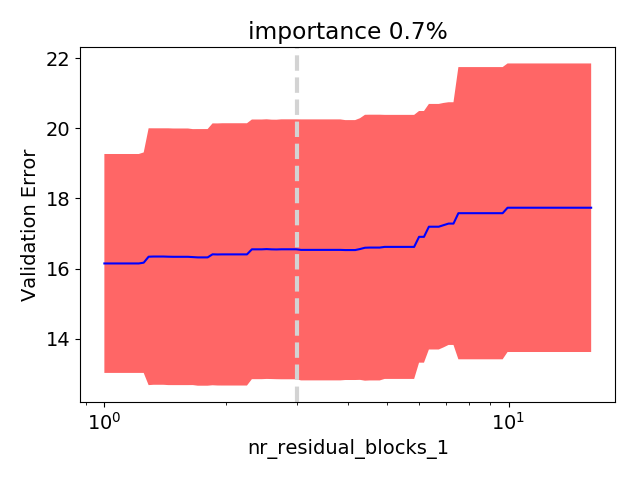}
	\includegraphics[width=0.30\linewidth]{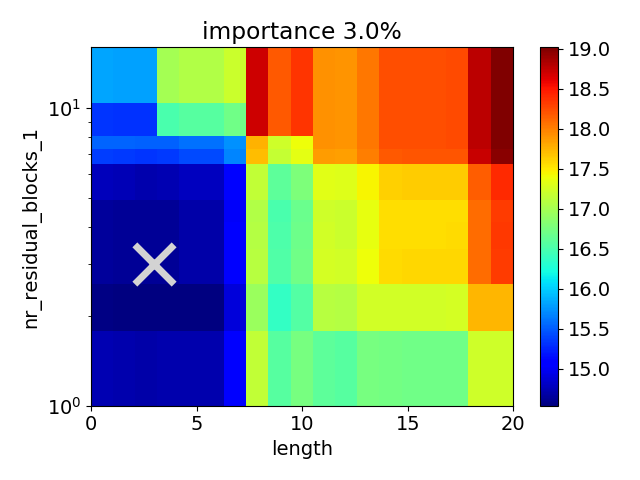}

	\caption{
		Parameter importance plots for three hyperparameters for training 400s (top row) and 1h (bottom row).
		The importance indicates the fraction of the variance explained by the individual choice(s).
		The value of the best found configuration on this budget is indicated by the dashed line/ gray x.
		While the learning rate remains the same throughout the budgets, other choices are heavily influenced by the constrained training time.
		For example, the characteristics for the number of residual blocks (middle) and the interaction between the number of residual blocks and the CutOut lengths (right), as well as their importance, change with the budget.
	}
	\label{fig:fanova_examples}
\end{figure}

We can observe different behavior for the various choices:
The learning rate remains important throughout all budgets and its value in the best found configuration only changes slightly.
On the other hand, some architectural parameters that affect the number of network parameters, e.g., the number of residual blocks within a block, can be heavily influenced by a constrained budget; this is, e.g., demonstrated in the plots in the middle of Fig. \ref{fig:fanova_examples}.
For the 400s training budget, the networks with less residuals blocks in the first block perform better on average and the importance of this choice is quite high.
With more training time, this choice becomes rather unimportant, and good configurations can be found with almost any value.
For the CutOut length and the number of residual blocks in the first block we find an interesting pattern (see plots on the right of Fig. \ref{fig:fanova_examples}).
While the good range for the CutOut length on the 400s budget is quite large (as long as the number of residual blocks is small enough), the picture changes for the larger budget.
There, the CutOut length becomes more restricted, and the number of blocks becomes less relevant.

Overall, from this analysis, we conclude that there is a close interaction between good architectural choices and hyperparameters on the one hand, and the runtime budget on the other hand. The common practice of optimizing just on the smallest budget and evaluating on the largest budget would be very wasteful in this case.

\section{Conclusions}

We showed that it is desirable and feasible to optimize neural architectures and hyperparameters jointly.
We also demonstrated the potentially strong effects that short training exerts on both architectural choices and hyperparameters, resulting in a poor correlation between the performance after short and long training periods, and showed how to sidestep this effect with incremental budget increases in the optimization process as implemented by the recent BOHB approach.

\clearpage

\section*{Acknowledgements}
This work has partly been supported by the European Research Council (ERC) under the European Union’s Horizon 2020 research and innovation programme under grant no.\ 716721. The authors acknowledge support by the High Performance and Cloud Computing Group at the Zentrum für Datenverarbeitung of the University of T{\"u}bingen, the state of Baden-W{\"u}rttemberg through bwHPC and the German Research Foundation (DFG) through grant no INST 37/935-1 FUGG.

{\small
\bibliography{ms}
\bibliographystyle{unsrtnat}
}

\clearpage

\appendix

\section{Joint Archicture and Hyperparameter Search Space}\label{sec:search_space}
Table \ref{tab:search_space} summarizes the 10 architectural choices and 7 hyperparameters we used, along with specific ranges and values of the best performing found configuration on the 3h budget. 

\begin{table}[ht]
  \centering
  \caption{The configuration space for joint architecture and hyperparameter optimization. Each numbered Residual Block, Residual Branches and Widen Factor correspond to each of the 3 main blocks of the architecture. The last column represents the configuration which performs the best after optimizing the search space with BOHB.}
  \begin{tabular}{|c|c|c|c|}
    \hline
	Hyperparameter & Range & Log-transform & Value\\
	\hline
	Initial Learning Rate & $[0.001, 1.0]$ & yes & $0.648188$\\
	Batch Size & $[32, 128]$ & yes & $89$\\
	$L_2$ regularization & $[0.00001, 0.001]$ & yes & $0.000339$\\
	Momentum & $[0.001, 0.99]$ & no & $0.099601$ \\
	MixUp $\alpha$ & $[0.0, 1.0]$ & no & $0.492042$ \\
	CutOut \textit{length} & $[0, 20]$ & no & $3$ \\
	ShakeDrop \textit{death rate} & $[0.0, 1.0]$ & no & $0.038439$ \\
    \hline
    $ResBlocks_1$ & $[1, 16]$ & yes & $3$ \\
    $ResBlocks_2$ & $[1, 16]$ & yes & $4$ \\
    $ResBlocks_3$ & $[1, 16]$ & yes & $2$ \\
    $ResBranches_1$ & $[1, 5]$ & no & $1$ \\
    $ResBranches_2$ & $[1, 5]$ & no & $1$ \\
    $ResBranches_3$ & $[1, 5]$ & no & $4$ \\
    $Filters_0$ & $[8, 32]$ & yes & $16$ \\
	$WidenFactor_1$ & $[0.5, 8.0]$ & yes & $6.241141$ \\
    $WidenFactor_2$ & $[0.5, 8.0]$ & yes & $1.388867$ \\
    $WidenFactor_3$ & $[0.5, 8.0]$ & yes & $3.344766$ \\
    \hline
  \end{tabular}
  \label{tab:search_space}
\end{table}

\section{Training details}\label{sec:training_details}

We use the PreAct ResNet-18~\citep{he-cvpr16} and WideResNet-28-10~\citep{zagoruy-corr16} architectures with Shake-Shake regularization (S-S-I; see \citet{gastaldi-iclr17} for details).
All networks were trained on one Nvidia GTX 1080Ti GPU.
The parameters of each model are initialized as described by~\citet{he-cvpr16} and trained using SGD with an initial learning rate of 0.1 and  with Nesterov's momentum of 0.9.
All the models in \ref{tab:config_budget} are trained for 3h with the initial learning rate annealed using a cosine function with $T_0 = 720s$, $T_{mult} = 2$ \citep{loshchilov-ICLR17SGDR}.
Furthermore, we also apply $L_2$ regularization with a factor of $10^{-4}$ and $5\cdot 10^{-4}$ to 3-branch and 2-branch networks, respectively.

During optimization, we split the CIFAR-10~\citep{krizhevsky-tech09a} datasets into a 45k data points for training, 5k for validation and a 10k for testing and we normalized the features by the per-channel mean and standard deviation.
For the training set we used standard data augmentation scheme used for CIFAR-10~\citep{he-cvpr16}, i.e. we first padded each image by 4 pixels, cropped a random 32x32 patch and flipped 50\% horizontally. Furthermore, we applied CutOut~\citep{cutout} with a mask length of 16 and MixUp~\citep{mixup} with an $\alpha$ value of $0.2$.

\section{Correlation across budgets}\label{sec:correlation_budgets}

\begin{figure}[ht]
	\centering
	\includegraphics[width=0.30\linewidth]{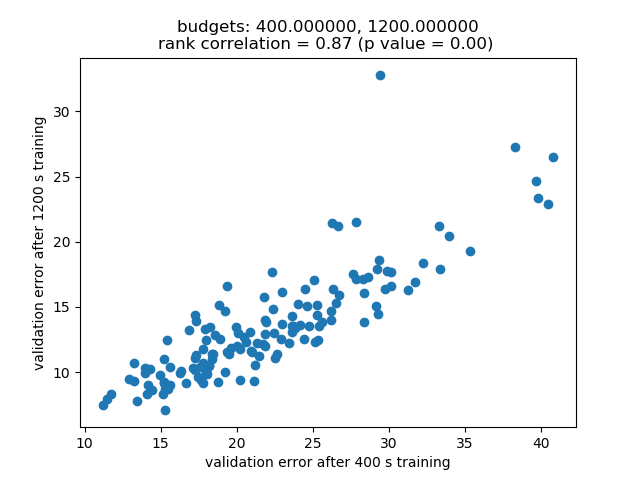}
	\includegraphics[width=0.30\linewidth]{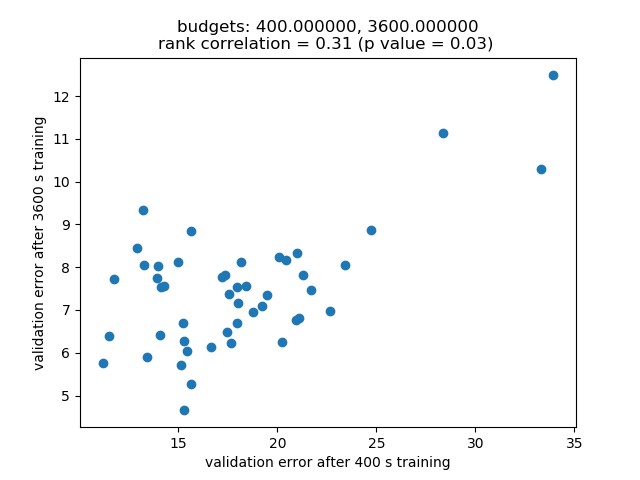}
	\includegraphics[width=0.30\linewidth]{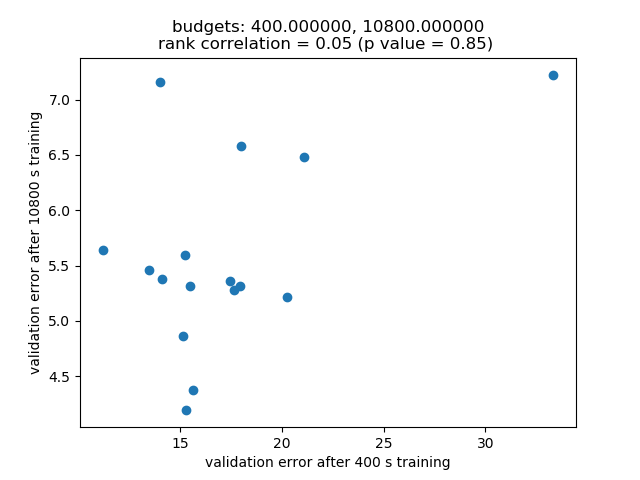}
	\includegraphics[width=0.30\linewidth]{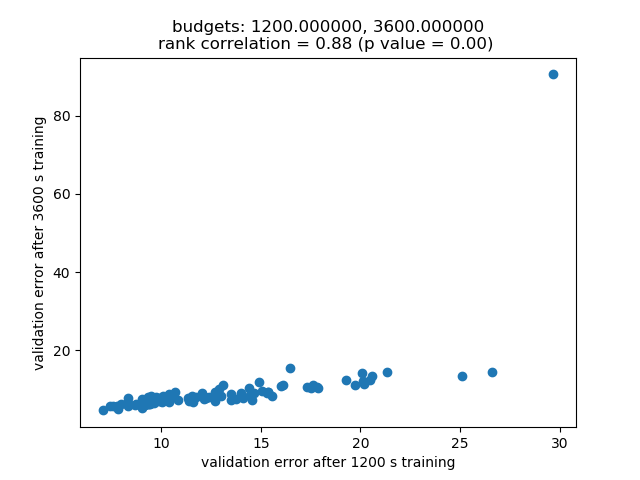}
	\includegraphics[width=0.30\linewidth]{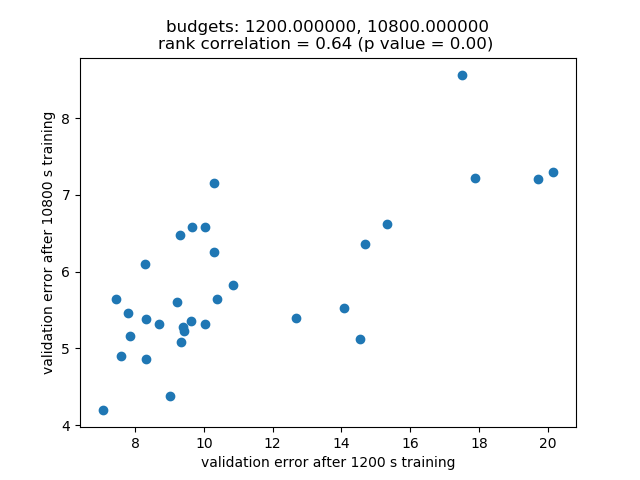}
	\includegraphics[width=0.30\linewidth]{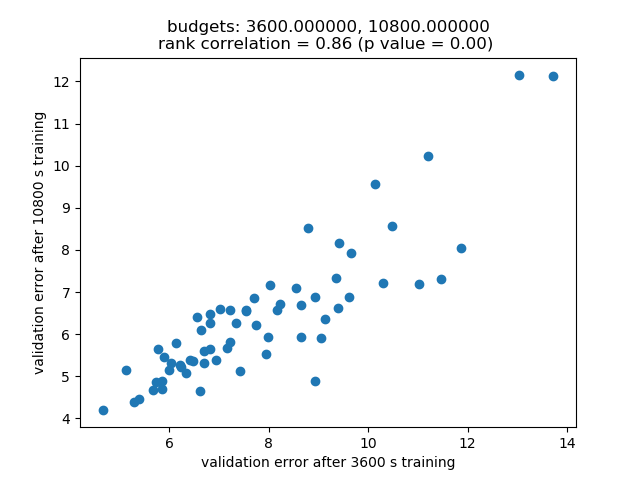}
	\caption{Correlation between the validation losses of different budgets and the largest budget of three hours.
	Due to the optimizer used, the collected data is heavily skewed towards the best performing configurations. 
	The rank correlation between the errors on different budgets shrinks as the difference in budget grows making the short budget uninformative about the best configurations for the longest training time.}
	\label{fig:loss_budget_correlation}
\end{figure}

Here we show the correlations between the different budgets presented in Section 4.3 in more detail.
Figure \ref{fig:loss_budget_correlation} shows all possible combinations of budgets.
One can clearly see that the correlation between adjacent budgets is very high, while it quickly degrades for larger differences.
Here, there is no correlation between the error rates on the smallest and the largest budget.
We want to emphasize again that these configurations are biased towards the best one on the largest budget (due to the selection during the optimization).
There is very likely a stronger correlation between these two budgets if worse configurations would be included.
That being said, our runs suggest that one cannot not solely optimize on a budget as small as 400 seconds and expect the best performance, if the final evaluation takes place on a budget 27 times larger.

\end{document}